\documentclass[11pt,a4paper]{article}
\usepackage[hyperref]{emnlp2020}
\usepackage{times}
\usepackage{latexsym}

\usepackage{microtype}

\usepackage{makecell}
\usepackage{graphicx}
\aclfinalcopy 
\def\aclpaperid{508} 

\usepackage{array}
\newcolumntype{C}[1]{>{\centering\arraybackslash\hspace{0pt}}p{#1}}

\newcommand{\wcds}{CE2.3k}
\newcommand{\sbcwcgpt}{GPT-Rank-CE}
\newcommand{\sbcds}{Rank-30k}
\newcommand{\sbcgpt}{GPT-Rank}
\newcommand{\lxnxds}{LN55k}
\newcommand{\lxnxgpt}{GPT-LN}
\newcommand{\allgpt}{GPT-ALL}
\newcommand{\sectionRef}[1]{\S \ref{#1}}

\title{The workweek is the best time to start a family -- A Study of GPT-2 Based Claim Generation}

\author {
 Shai Gretz\thanks{\ \ These authors equally contributed to this work.} , Yonatan Bilu\footnotemark[1] ,
 Edo Cohen-Karlik and Noam Slonim\\
 IBM Research\\
 \{avishaig,yonatanb,noams\}@il.ibm.com\\
\{edo.cohen\}@ibm.com
}

\date{}

\begin{document}
\maketitle
\begin{abstract}
Argument generation is a challenging task whose research is timely considering its potential impact on social media and the dissemination of information. Here we suggest a pipeline based on GPT-2 for generating coherent claims, and explore the types of claims that it produces, and their veracity, using an array of manual and automatic assessments. In addition, we explore the interplay between this task and the task of Claim Retrieval, 
showing how they can complement one another.     
\end{abstract}

\section{Introduction}\label{sec:intro}
Argument Mining had traditionally focused on the detection and retrieval of arguments, and the classification of their types and of the relations among them. Recently, there has been growing interest in argument synthesis. Here we suggest a pipeline for addressing this task relying on the GPT-2 language model \cite{radford2019language}, examine how it can be enhanced to provide better arguments, and analyze the types of arguments being produced.
Specifically, we are interested in 
\textit{Claim Generation}, where the input is a debate topic, phrased as a proposed policy, and the output is a concise assertion with a clear stance on this topic. 

We start by fine-tuning GPT-2 on a collection of topics and associated claims. Since several such datasets are available, we examine which of them tend to yield better claims, and observe that merging all such sources together does not necessarily yield better results.
In addition, we explore two ways in which context can be 
added to the generation process, beyond providing the topic itself: 
(i) framing the topic with the first sentence from its corresponding Wikipedia page; and (ii) framing the claim by directing it to consider a specific aspect. We find that the former can improve the generated output, but the latter does not -- 
at least in the way it is done here.
Following \citet{bilu-slonim-2016-claim}, we also examine a post-generation ranking 
step that aims to select the correctly generated claims. We find that existing \textit{Claim Detection} tools can 
serve as a filter to 
significantly enhance generation quality. 


Our evaluation incorporates automatic measures and manual labeling. Specifically, we introduce an annotation task aiming to assess the \textit{plausibility} of generated claims, i.e., to what degree is it plausible that a human will make it. 
We report results on a 
test set of $96$ topics, 
demonstrating the validity of our approach to topics not seen in training or development.
In addition, we manually annotate the generated claims for whether they are factual claims, or opinion based, and further aim to assess whether the former represent true facts. 

Finally, we observe that manually labeled datasets used to fine-tune GPT-2 are not essential, and that relying on the output of a \textit{Claim Retrieval}\footnote{Given a topic of interest, Claim Retrieval is the task of retrieving relevant claims from a corpus; Claim Detection is the task of determining whether a given text is a relevant claim.} 
engine for this fine-tuning, may suffice.
In addition, we compare the generated claims to an existing large-scale collection of claims for the same topics, and conclude that the generated claims tend to be novel, and hence may augment traditional Argument Mining techniques in automatically providing claims for a given topic.

Henceforth, we denote the initial output of GPT-2 for a given prompt as \textit{generated text (GT)}. Thus, our task is to define a process by which as many of the GTs as possible will represent claims that are relevant to the provided prompt. 
\section{Related Work}\label{sec:related}
In classical Natural Language Generation (NLG) tasks -- Machine Translation, Summarization, and Question Answering -- the semantic content of the output strongly depends on the input.
Argument Generation, alongside Story Generation \cite{fan-etal-2018-hierarchical}, occupies a parallel venue, where the output should satisfy stylistic and rhetorical constraints -- yet no well-defined semantic goal -- with much room and desire for innovation.

Approaches to argument generation have included traditional NLG architectures \cite{zukerman1998bayesian, carenini2006generating}; assembling arguments from given, smaller argumentative units \cite{walton2012carneades, reisert2015computational, wachsmuth-etal-2018-argumentation, el-baff-etal-2019-computational}; welding the topic of the debate to appropriate predicates \cite{bilu-slonim-2016-claim}; and using predefined argument templates \cite{bilu2019argument}. Of particular interest is the generation of counter arguments, for which solutions include an encoder-decoder architecture \cite{hidey-mckeown-2019-fixed}, which may be augmented by a retrieval system \cite{hua-etal-2019-argument-generation, hua-wang-2018-neural}, or alternatively offering ``general purpose" rebuttal based on similarity to predefined claims \cite{orbach-etal-2019-dataset}.

Concurrent with our work, and most similar, is \citet{schiller2020aspect}, who frame the Aspect-Controlled Argument Generation problem as follows - given a topic, a stance and an aspect, generate an argument with the given stance towards the topic, which discusses the given aspect. They fine-tune CTRL \cite{keskar2019ctrl} over claims from $8$ controversial topics, and mostly use automatic measures to assess claim generation over the same 
$8$ topics. 
By contrast, here we are interested in a less restricted setting and explore the properties of the generated claims. Specifically, we fine-tune GPT-2 on claims coming from diverse sets of $71$-$192$ topics, and evaluate claims generated for $96$ novel topics. 

In this work, we assess the contribution of context to the quality of generated claims. In \citet{durmus-etal-2019-role}, context is defined as the path from a thesis (topic) node to a leaf (claim) node in an argument tree. In this work, however, we consider only arguments of depth 1, directly addressing the topic, and leave context of larger depth to future work.

Additionally, for development and evaluation we use human annotations alongside automatic measures, aiming to answer nuanced questions - is it plausible that the claims be asserted by a human? do the generated claims tend to be opinions or factual? and, when they are the latter, do they tend to be factually true?

\section{Experimental Details}\label{sec:experimental_details}

\subsection{Data}
\label{sec:data}

We compare the performance of fine-tuning GPT-2 on three argument datasets, two publicly available and one proprietary. 

\textbf{\sbcds{}}. This dataset includes $30k$ arguments for $71$ topics, labeled for their quality \cite{gretz2019largescale}. For fine-tuning GPT-2 we consider all arguments with quality score (denoted there as WA-score) $> 0.9$, resulting in $10{,}669$ arguments. These arguments are typically $1$-$2$ sentences long.

\textbf{\wcds{}}. This dataset consists of $2.3k$ manually curated claims extracted from Wikipedia for $58$ topics \cite{rinott-etal-2015-show}. 
These claims are usually sub-sentence, concise phrases. We exclude claims for topics which are part of our dev set (see below). Further, we ``wikify" each topic, i.e., automatically map each topic to a corresponding Wikipedia title \cite{shnayderman2019fast}, and remove topics for which no such mapping is found. After this filtering, we remain with $1{,}489$ claims for $29$ topics.

\textbf{\lxnxds{}}. This proprietary dataset consists of $55{,}024$ manually curated claims for the $192$ topics in the train set of \citet{EinDor2019CorpusWA}. These claims were extracted from a  corpus  of  some  $400$ million  newspaper articles  provided  by LexisNexis,\footnote{\url{https://www.lexisnexis.com/en-us/home.page}} as done in \citet{EinDor2019CorpusWA} for evidence rather than claims.

Whereas fine-tuning is done on varied data-sources, for evaluation we focus on the dev and test topics from \citet{EinDor2019CorpusWA}. We exclude from both sets topics that are present in the \sbcds{} dataset, resulting in a dev set of $35$ topics and test set of $96$ topics (see Appendix).

Throughout this work, we consider debatable topics which correspond to a single Wikipedia title, phrased as a suggestion for a policy -- e.g., \textit{We should increase the use of telemedicine}, or as a valuation analysis -- e.g., \textit{telemedicine brings 
more harm than good}.

\subsection{Model}

For all experiments we fine-tune the medium-size GPT-2-355M model \cite{radford2019language}, utilizing the gpt-2-simple library.\footnote{\url{https://github.com/minimaxir/gpt-2-simple}} In order for the model to condition on topics, we represent each (\textit{topic}, \textit{claim}) pair from the training data as a single sequence, separated by a delimiter. In generation, the model is provided with a prompt in the form of a topic followed by a delimiter. We used \textit{top-k} truncation with $k=40$ and a conservative temperature of $0.7$, to accommodate a more readable, coherent output, while maintaining a level of creativity. We leave exploring other sampling techniques (e.g., \citet{holtzman2019curious}) to future work. We restricted the length of each generated text to $50$ BPE tokens, as preliminary experiments showed that very few GTs were longer. In addition, GTs were cleaned by removing non-ascii characters, parenthesis, single quotation marks, and some other erroneous symbols.

\subsection{Automatic Evaluation}
\label{sec:automaticEvaluation}

For evaluation, we consider perplexity and prefix ranking accuracy \cite{fan-etal-2018-hierarchical}, considering the claims extracted by  \citet{ajjour-etal-2019-modeling} 
alongside their listed topics.
\footnote{This dataset contains $12{,}326$ claims from $465$ topics extracted from \url{debatepedia.org}. We rephrase topics therein to fit our phrasing by adding the text ``We should support" before of the listed topic.} For prefix ranking accuracy we condition each such claim on its real topic, as well as on $9$ other random topics, and compute the fraction of times where conditioning on the real topic yields the highest probability by the fine-tuned model. For both evaluation measures, we report statistics for $10$ samples of $100$ claims sampled uniformly. 
Importantly, this dataset is independent of all the ones examined here, and so presumably not biased in favor of any of them. Due to the difference in style and topics from the training sets, the fine-tuned models may exhibit high perplexity, so it should be taken as a comparative measure, rather than an absolute one.

In addition, we evaluate the GTs by their \textit{quality} and \textit{stance} scores. For obtaining a quality score, we fine-tune BERT \cite{devlin2018bert} on \sbcds{}, as 
in \citet{gretz2019largescale}. This score aims to capture how well the output is written, giving preference to grammar, clarity and correct spelling. For obtaining a stance score, we utilize a proprietary internal service, based on a BERT model fine-tuned over the \lxnxds{} claims which were manually labeled for stance \cite{bar-haim-etal-2017-stance}. A positive score indicates that a claim supports the topic, a negative score that it contests it, while a score close to zero suggests no clear stance. Since we are only interested in whether or not a sentence has a clear stance, we take the absolute value of the score.
For both scores, we report statistics for $10$ samples of $100$ GTs sampled uniformly from the respective set.

\subsection{Annotation Tasks}
\label{sec:annotationTasks}
To further assess the quality of GTs we annotate their \textit{plausibility} and \textit{stance}.  We do this in a cascade -- only GTs considered plausible are subsequently annotated for their stance. The motivation for these two tasks is that together they enable us to assess the ``claimness" of GTs, i.e., to determine to what extent the GTs represent coherent claims, relevant to the given topic. We used the Appen crowd-sourcing platform,\footnote{\url{www.appen.com}} with $7$ annotators to annotate each GT. To control for annotation quality, we included hidden test questions, comprised of previously annotated rows with high confidence. Annotations by annotators with low accuracy on the test questions were removed (below $75\%$ for plausibility and $80\%$ for stance). 
Further, we relied on a channel of annotators which performed well on previous related tasks. For each task, we report inter-annotator agreement defined as the average Cohen's Kappa of annotators which have at least $50$ common judgements with at least $5$ other annotators.

\textbf{Plausibility}. In this task, given the GT only, without the context of its respective topic, the annotator should determine if it is plausible that a human would make this claim, considering grammar, coherence, and general ``common sense". This task can be considered an extension of the \textit{readability} task that is usually used to evaluate the quality of generated text (e.g., \citet{beers2009syntactic}), while further asking to utilize common knowledge to judge that the content itself makes sense. For example, in the GT \textit{making blood donation free will help promote primary care}, the notion of \textit{making blood donation free} does not make sense as it is a voluntary act, hence the GT should be deemed implausible. 
A GT is considered plausible if $\geq 70\%$ of the annotators considered it as such. The average inter-annotator Cohen's Kappa obtained in this task is $0.37$, which is common for such a subjective task (see, e.g., \citet{EinDor2019CorpusWA} and \citet{Boltuzic2014BackUY}).

\textbf{Stance}. In this task we presented the annotators with GTs that were considered plausible, together with their respective topics. Annotators were asked to determine if the text \textit{supports} the topic, \textit{contests} it, or does not have a stance towards it. The label of the GT is determined by the majority vote, and if there is no majority label, it is considered as having no stance. As in the automatic measure of stance, we are mainly interested in evaluating if a GT bears \textit{any} stance towards the topic, thus we consider both \textit{supports} and \textit{contests} labels as positives when reporting stance. The average inter-annotator Cohen's Kappa obtained in this task is $0.81$.

Table \ref{table:exampleClaims} shows examples of three types of labeled GTs -- plausible and stance-bearing; plausible with no stance; and implausible. The results of these annotation tasks 
are made available as part of this work.\footnote{\url{https://www.research.ibm.com/haifa/dept/vst/debating\_data.shtml}} The complete annotation 
guidelines are shared in 
the Appendix.

\section{Initial Generation}\label{sec:initial_gen}

Our first question was to examine the impact of the data used for fine-tuning GPT-2, aiming to identify an effective model that relies on publicly available data, and a presumably superior one that further relies on proprietary data of a much larger size.




\noindent
\textbf{Publicly available data}. We considered \sbcds{} alone, and combined with \wcds{}. We fine-tuned GPT-2 for $2k$ steps on the former, and $4k$ steps on the latter. We denote the obtained 
models \sbcgpt{} and \sbcwcgpt{}, respectively.

\noindent
\textbf{Proprietary data}. We considered \lxnxds{} alone, as well as combined with all publicly available data.
We fine-tuned GPT-2 for $8k$ steps on both. We denote the obtained 
models \lxnxgpt{} and \allgpt{}, respectively.\footnote{In section \ref{sec:retrieved}, we describe the retrieval of $4.5k$ (ostensible) claims from Wikipedia using a proprietary Claim Retrieval server. These claims are included in \allgpt{}.}

For each of the $4$ models we generated a total of $175$ GTs, $5$ conditioned on each of the $35$ dev topics. Note that the models are fine-tuned on datasets containing both supporting and contesting arguments, thus they may generate GTs of both stances as well.
The manual and automatic evaluation of these GTs is described next. 


\begin{table*}[t]
\small
\begin{center}
\begin{tabular}{ |c|c|c|c|c|c|c|c|c|  }
\hline
                                & PL & PL + ST  & PPL & PR & P-QU & P-ST & P-QU* & P-ST* \\
\hline
\lxnxgpt{}     & $75.4\%$      & $\textbf{68\%}$ & $188.9$  & $0.69$  & $0.75$ & $0.99$ & $0.78$ & $1.00$         \\
\hline
\allgpt{}{} & $78.9\%$      & $62.3\%$ & $82.7$       & $0.74$ & $0.76$ & $0.97$ & $0.79$ & $1.00$         \\
\Xhline{2\arrayrulewidth}
\sbcgpt{}      & $53.1\%$       & $51.4\%$   & $150.8$      & $0.75$  & $0.85$ & $0.99$ & $0.85$ & $1.00$         \\
\hline
\sbcwcgpt{}  & $64.6\%$      & $\textbf{54.9\%}$  & $388.4$       & $0.65$ & $0.8$ & $0.98$ & $0.84$ & $1.00$ \\
\hline
 \end{tabular}
 \end{center}
 \caption{Results on the dev set of models fine-tuned on proprietary (top 2) and publicly available (bottom 2) data sources. PL = ratio of plausible claims, PL + ST = ratio of plausible and stance bearing claims, PPL = perplexity, PR = prompt ranking accuracy, P-QU = predicted quality, P-ST = predicted (absolute) stance. Asterisk indicates values for the training set.}
\label{table:resultsInitial}
\end{table*}

As seen in Table \ref{table:resultsInitial} both proprietary models -- fine-tuned on much larger datasets -- yield more plausible and stance-bearing GTs than their counterparts.

Among the proprietary-based models, while \allgpt{} has an advantage in plausibility, perplexity, and prefix ranking accuracy, \lxnxgpt{} is better when considering the ratio of GTs which are both plausible and stance-bearing - with $68\%$ ($119/175$) such GTs, compared to $62.3\%$ ($109/175$) for GPT-2-ALL. It seems that adding more data, varied in type and style, could negatively impact the relevance and usefulness of GTs. Thus, we choose \lxnxgpt{} as the model to utilize for subsequent experiments.

As for the publicly-based models, \sbcwcgpt{} has a small advantage in plausible and stance-bearing GTs, compared to \sbcgpt{}. However, the performance of the latter is typically much better in the automatic measures. Especially, we note the advantage in predicted quality - as expected, generated arguments from the \sbcgpt{} model have higher quality, as both this model and the argument quality model were trained on a similar type of data. However, when adding the \wcds{} dataset to the training set, the quality of GTs declines. Thus, even though the differences between the two models are overall not substantial, we choose \sbcgpt{} for subsequent experiments.

\begin{table*}[t]
\small
\begin{center}
\begin{tabular}{ |p{2.5cm}|p{7.5cm}|p{1.5cm}|p{2cm}|  }
\hline
\textbf{Topic} & \textbf{GT} & \textbf{Model}  & \textbf{Label}  \\
\hline
We should abandon democracy & A proper democracy is good for the country  & \lxnxgpt{} & Plausible and has stance\\
\hline
We should lower the drinking age & the age of majority in the country was lowered to 18 &  \lxnxgpt{} & Plausible with no stance\\
\hline
We should ban free newspapers & free newspapers reduce crime  & \lxnxgpt{} & Implausible\\
\hline

We should increase government regulation & we need regulation to make sure our country is protected. with more government involvement in our daily lives, businesses can hire more workers and produce more output. & \sbcgpt{} & Plausible and has stance\\
\hline
We should fight for Palestinian independence & the liberation of Palestine will be impossible if the Palestinians are ruled by corrupt Israeli and Palestinian governments & \sbcgpt{} & Plausible with no stance\\
\hline
We should ban lotteries & lotteries are a great way for children to learn about different cultures and find similar things to do & \sbcgpt{} & Implausible\\
\hline
 \end{tabular}
 \end{center}
 \caption{Examples of GTs generated by the \lxnxgpt{} and \sbcgpt{} models, labeled for plausibility and stance.}
\label{table:exampleClaims}
\end{table*}


It should be noted that there is a clear difference between the GTs of \lxnxgpt{} and \sbcgpt{}, as evident in Table \ref{table:exampleClaims}.
The former are short ($12.4$ tokens on average), and may contain utterances with as few as $3$-$4$ tokens (as in the GT in row 3). 
By contrast, GTs generated by \sbcgpt{} contain $23$ tokens on average, and $22/175$ of them contain at least two sentences (as in the GT in row 4).
In addition, shorter GTs tend to be plausible - on average, plausible GTs from \lxnxgpt{} have $12.1$ tokens, compared to $15.4$ tokens for implausible GTs. Likewise, plausible GTs from \sbcgpt{} contain $20.5$ tokens, on average, compared to $26$ tokens for implausible GTs.

We note that for all models, the predicted quality and stance strength are only slightly lower than their counterpart measures on the training set, suggesting that generation tends to maintain these values.

\section{Adding context}\label{sec:context}

Can we improve GTs by conditioning their generation on more context? To evaluate this hypothesis we considered two context variations, one in which we frame the topic and the other in which we frame the claim.

\textbf{Framing the topic}. We prepend to the topic the first sentence from the Wikipedia page describing the topic, to explore whether this added knowledge could guide models to generate more relevant and meaningful GTs. The motivation for selecting the first sentence from Wikipedia is to provide the model a concise guidance towards the respective topic via the main terms it may relate to, which usually appear in the first Wikipedia sentence. The relevant Wikipedia page is found by Wikifying the topic, as described in \sectionRef{sec:data}.

\textbf{Framing the claim}. We also tried to  append to the topic a short sentence describing an aspect relevant to discussing it, hypothesizing that adding a concrete aspect will guide the generation process in that direction. Unfortunately, this did not work well, and details are deferred to the appendix.

\paragraph{Evaluation:} We fine-tune GPT-2 from scratch on the modified training data of \sbcds{} and \lxnxds{} and refer to the new models as \textit{\sbcgpt{}-FWS}, \textit{\lxnxgpt{}-FWS} (First Wikipedia Sentence, when framing the topic). We generate a sample of 5 (\textit{\sbcgpt{}-FWS}) or 10 (\textit{\lxnxgpt{}-FWS}) GTs per dev topic.




\begin{table*}[t]
\small
\begin{center}
\begin{tabular}{ |c|c|c|c|c|c|c|  }
\hline
                                & PL & PL + ST  & PPL & PR & P-QU & P-ST \\
\hline
\lxnxgpt{}     & $76.6\%$      & $66.3\%$ & $188.9$      & $0.69$  & $0.75$ & $0.99$         \\
\hline
\lxnxgpt{}-FWS & $85.1\%$      & $\textbf{73.1\%}$ & $88.6$       & $0.74$ & $0.76$ & $0.99$         \\
\Xhline{2\arrayrulewidth}
\sbcgpt{}      & $53.1\%$       & $\textbf{51.4\%}$   & $150.8$      & $0.75$  & $0.85$ & $0.99$         \\
\hline
\sbcgpt{}-FWS  & $58.9\%$      & $49.7\%$  & $71.6$       & $0.76$ & $0.83$ & $0.97$ \\
\hline
 \end{tabular}
 \end{center}
 \caption{Results on the dev set of models with and without conditioning on the first sentence of the Wikipedia page corresponding to the topic. Column titles as in Table \ref{table:resultsInitial}. For \sbcgpt{} we used $175$ GTs as per Section \ref{sec:initial_gen}. For \lxnxgpt{}, data includes an additional $175$ GTs. Hence, numbers here differ from Table \ref{table:resultsInitial}. }
\label{table:resultsContext}
\end{table*}


\paragraph{Results:}  Table \ref{table:resultsContext} presents the results for the FWS models. For both FWS models the perplexity has improved, as well as the plausibility of GTs, presumably, since the added context helps to avoid some illogical phrases. For example, the GT \textit{The human condition is the greatest human achievement} for the topic \textit{We should subsidize the human mission to Mars} which was generated by \lxnxgpt{} was considered implausible, whereas all GTs for this topic generated by \lxnxgpt{}-FWS were considered plausible. After stance labeling, the advantage of \lxnxgpt{}-FWS remains, while \sbcgpt{}-FWS performs slightly worse. In addition, the \sbcgpt{}-FWS is slightly worse in predicted quality and stance. Thus, for further experiments, we chose the \lxnxgpt{}-FWS and \sbcgpt{} models.
\section{Factual, Opinion, and Generic Claims}

An interesting facet when considering argumentative claims, is whether they attempt to convey facts, or rather personal opinions. Thus, we explored if GTs generated by our two models are characterized as more factual or opinionated. Further, given growing concern over misuse of language models such as GPT-2 to spread fake news and misinformation \cite{NIPS2019_9106,solaiman2019release}, we assessed the truth value of GTs deemed factual.
For this purpose, we first sampled $200$ plausible and stance-bearing GTs each generated by \lxnxgpt{}-FWS and \sbcgpt{}, respectively, and annotated all $400$ GTs for being an opinion or (ostensibly) factual, using the Appen platform, and relying on similar annotation controls as described in \sectionRef{sec:annotationTasks}. 
The results of this annotation task are made available as part of this work, and the annotation guidelines are shared in the Appendix. 
The average inter-annotator agreement was $0.25$.

When considering labels with a majority vote of at least $70\%$, $70$ of the GTs generated by \sbcgpt{} are considered factual and $63$ opinion, as opposed to $46$ and $105$ of those generated by \lxnxgpt{}-FWS, respectively. A possible explanation is that 
\sbcds{} claims -- on which \sbcgpt{} was fine-tuned -- tend to be more elaborate and explanatory, describing a cause and effect that correspondingly yields more factual GTs; e.g., the GT \textit{genetic engineering can help further scientific developments in cancer treatment, as well as improve the long term prognosis of such diseases as help 
maintain a safe and effective regulatory regime for their development}, for the topic \textit{We should further exploit genetic engineering}.
By contrast, \lxnxds{} claims are often short and concise, and perhaps more prone to express the journalist opinion; hence, training on these data yields more opinionated GTs,
e.g., \textit{the ``sex" revolution has failed} or \textit{the gender pay gap is unfair}. Indeed, the average number of tokens in factual GTs is $17.3$, compared to $14.2$ for opinion GTs.

Next, we aimed to assess whether factual GTs are indeed true. A random sample 
of $23$ and $40$ factual GTs generated by \lxnxgpt{}-FWS and \sbcgpt{}, respectively, were labeled for their truth value by a professional debater experienced in this task, that also was asked to assess whether the ``fake facts" were nonetheless common in contemporary discourse.

Of the $23$ \lxnxgpt{}-FWS GTs, $13$ were considered true, the others being a mix of false or non-factual GTs. 
The true GTs include some simple, almost trivial statements such as \textit{Speed limits are designed to help reduce road fatalities}, or more evidence-based facts such as \textit{rat poisons have been linked to the development of Parkinson's disease, Alzheimer's disease and migraines}. Among the $4$ false GTs, it is interesting, albeit perhaps unsurprising, to find that $2$ were marked as common in discourse: \textit{Flu vaccinations are associated with higher rates of adverse drug reactions and serious health complications}, and \textit{poly-amorous relationships are linked to higher levels of sexual risk}.

For the $40$ \sbcgpt{} factual GTs, $21$ were deemed true. 
Overall, the ratio of true GTs is similar to that of \lxnxgpt{}-FWS GTs. It seems that some of the other GTs are mixed, characterized by opening with an opinionated statement, which is followed by a factual claim, e.g., \textit{we should not abandon chain stores} (Opinion) \textit{as they provide a steady supply of goods and services to the community} (True fact). One of the $3$ false GTs could be considered common in discourse, \textit{the alternative vote would cause voters to be disenfranchised}.

The aforementioned short GTs suggested that GTs tend to be rather generic, in the sense that stating that 
something ``has failed" or ``is unfair", can be done (coherently) for a great variety of contexts. Indeed, such GTs are reminiscent of those generated by \citet{bilu-slonim-2016-claim}. To assess to what extent such GTs are generic, we sampled $100$ of them, and annotated them ourselves. In this sample, $54$ of the GTs were deemed generic, suggesting that such GTs are prevalent, but by no means the 
only types of texts being generated.

\section{The Complete Pipeline}\label{sec:classify}

\subsection{Ranking Generated Claims}
\label{sec:ranking}
So far we have assessed the overall ability of the models to generate relevant claims. A natural question is whether one can efficiently rank the obtained GTs, retaining only the most attractive ones for downstream tasks. This could be considered somewhat analogous to Claim Retrieval tasks, where first a large amount of argument candidates is retrieved, and are then ranked according to their relevance (e.g., \citet{levy-etal-2014-context,stab-etal-2018-cross,EinDor2019CorpusWA}).

We considered three existing models for ranking GTs - the argument quality and stance models described in \sectionRef{sec:automaticEvaluation}, and a Claim Detection (CD) proprietary service, obtained by training a BERT model on 
\lxnxds{}. The data for training the model is augmented with negative samples from the same corpus -- sub-sentential fragments which were labeled as non-claims. The objective of the model is to differentiate between claims and non-claims, and is similar to that described in \citet{EinDor2019CorpusWA} for Evidence detection.
For evaluation we considered GTs generated on the dev set by \sbcgpt{} and \lxnxgpt{}-FWS for which we had a definite label for relevance to the topic. Specifically, GTs which were annotated as ``implausible" by a majority of annotators were assigned a label of $0$. GTs which were annotated as plausible, and then annotated for stance, were labeled according to the latter annotation: $1$ if they were annotated as \textit{Pro} or \textit{Con}, and $0$ otherwise. In total, we considered $211$ positive and $120$ negative GTs.

Overall, the CD score is best correlated with the labels - Pearson's $\rho=0.41$, compared to $0.12$ for (absolute) stance, and $0.01$ for argument quality. In addition, we ranked the GTs within each topic w.r.t each score, and calculated the ratio between the number of positives in the top $3$ and bottom $3$. As before, CD is preferred, with $81/40$ positives in the top/bottom, compared to $70/56$ (stance) and $71/67$ (argument quality). See a short discussion about this result in the Appendix. 

Accordingly, we defined the generation pipeline as follows: (i) Fine-tune GPT-2 to obtain \sbcgpt{} (Model-1) or \lxnxgpt{}-FWS (Model-2); (ii) Generate with the topic as a prompt (Model-1), or prepend the -- automatically extracted -- first sentence of the associated Wikipedia article to the topic and use the resultant text as a prompt; (iii) rank the obtained GTs according to their CD score. 
In principle, one could set a strict threshold on the CD score, and generate a large number of texts until a sufficient number pass this threshold. We plan to investigate this direction in future work.

\subsection{Test Set Results}
With the above pipeline, we now proceed to generate $20$ GTs for each of the $96$ topics in the test set, using the \lxnxgpt{}-FWS and \sbcgpt{} models. We then take the top $7$ GTs according to the CD score, per topic, resulting in $672$ GTs overall for each model. As done for the dev set, we label these GTs for plausibility and stance, as well as calculate their predicted quality and stance.

\begin{table}[t]
\small
\begin{center}
\begin{tabular}{ |C{2.0cm}|c|c|c|c|  }
\hline
& PL & PL + ST  & P-QU & P-ST \\
\hline
\lxnxgpt{}-FWS & $82.4\%$ & $79.5\%$ & $0.78$ & $0.97$         \\
\Xhline{2\arrayrulewidth}
\sbcgpt{}      & $58.8\%$       & $57\%$   & $0.85$ & $0.98$         \\
\hline
 \end{tabular}
 \end{center}
 \caption{Results on the test set of the \lxnxgpt{}-FWS and \sbcgpt{} models, with ranking using the claim detection model. Column titles as in Table \ref{table:resultsInitial}.}
\label{table:resultsTest}
\end{table}
Results are presented in Table \ref{table:resultsTest}. The overall ratio of GTs perceived as both plausible and carrying stance for the \lxnxgpt{}-FWS model and the \sbcgpt{} model are $79.5\%$ and $57\%$, respectively, conveying the advantage of fine-tuning on much larger data (see the appendix for examples). In addition, our test set results echo the results obtained on the dev set, suggesting that our analysis on the dev set is relevant for the test set as well, and that our models generalize well to 
unseen topics.

\section{Claim Generation vs. Claim Retrieval}

Given a controversial topic, Claim Generation and Claim Retrieval both aim to provide claims pertaining to it. It is therefore interesting to understand the interplay between the two tasks. Specifically, thinking of Claim Generation as a mean to augment the output of Claim Retrieval, we ask whether GTs tend to be novel, or a repetition of retrieved claims, and how does the quality of the two compare. In addition, we explore how Claim Retrieval can facilitate the training of the Claim Generation pipeline suggested in this work.

\textbf{How novel are the generated claims?}
Similar to the manually-curated claims of the \lxnxds{} dataset, we also had access to such claims pertaining to $34/35$ topics in the dev set (henceforth, the LN claims). For comparison we used $169$ GTs ($5$ per topic, one duplicate removed) from the GTs generated by \lxnxgpt{} for these $34$ topics (see Section \sectionRef{sec:initial_gen}).
To measure similarity between GTs and LN claims we fine-tuned BERT on a Semantic Text Similarity benchmark \cite{cer2017semeval}. The resultant model was used to find for each GT the top matching LN claim. Manual examination suggests that a score of $0.75$ roughly differentiates pairs with semantically similar claims and those which are not (Table \ref{table:matchedClaims}). 
\begin{table*}[t]
\small
\begin{center}
\begin{tabular}{ |c|l|l|c|  }
\hline
& \textbf{Generated claim} & \textbf{Matched claim}  & \textbf{Score}   \\
\hline
1 & natural gas has positive effects on the environment &
natural gas can have a negative environmental effect & 0.85 \\
\hline
2 & alternative medicine could be a good option for  &
alternative medicine could be useful & 0.76 \\
& some patients & &\\
\hline
3 & the lottery could drive away investment & 
lottery could be a significant source of revenue & 0.75 \\
\hline
4 & lower retirement ages would promote more &
a higher minimum retirement age would lead to people & 0.74 \\
& long-term job stability &  working longer translating in greater economic output & \\
\hline
 \end{tabular}
 \end{center}
 \caption{Examples of matching of generated claims to manually-curated claims.}
\label{table:matchedClaims}
\end{table*}
Note that semantically similar claims may still have opposing stance, but in this case we also consider the GT as appearing in the corpus (in its negated form). 

Taking all pairs with score $\geq 0.75$, we get that only $20/169$ of the GTs have a semantically-similar counterpart among the LN claims, suggesting that GTs tend to be novel.
Moreover, we see that the match score is well correlated with the number of annotators who labeled a GT as plausible (Pearson's $\rho=0.31$) or as having a stance ($\rho=0.47$). Similarly, in general, $127/169$ GTs were determined by human annotators to be plausible and $114/169$ as having a stance. In comparison, $19/20$
GTs with match score $\geq 0.75$, were deemed both plausible and as having a stance. This suggests, as may be expected, that GTs are more likely to represent valid claims if they already appear in some phrasing within a human-authored corpus. Future work might use this to validate GTs, or, conversely, to guide claim retrieval.

\textbf{How good are the generated claims?}
Having matched GTs to ``real" claims allows us to compare not only their novelty, but also their quality. Namely, for each of the $169$ pairs we asked crowd annotators which of the two claims ``would have been preferred by most people to discuss the topic?", using the same process as in section \sectionRef{sec:experimental_details}. Among these pairs, in $41$ cases both claims appeared to be similarly good (a $3$:$4$ split); in $57$ the GT is preferred; and in $71$ the LN claim is considered better. Among the $20$ pairs which are highly similar, in $4$ both claims are equally good, in $13$ the GT is better and in $4$ the LN claim is preferred. Thus, at least in this small sample, when the two claims are conveying a similar message, human annotators seem to prefer the GPT-2 version over the human authored one.

\textbf{Can claim retrieval facilitate generation?}\label{sec:retrieved}
The suggested pipeline assumes access to a dataset of actual claims to fine-tune GPT-2. However, initial analysis suggest that even with no {\it a-priory\/} labeled data, having access to a high quality 
Claim Retrieval engine, can be enough to facilitate Claim Generation. Using a propriety Claim Retrieval server, we first query Wikipedia to retrieve sentence candidates, in a similar process to that described in \citet{EinDor2019CorpusWA} for retrieving Evidence candidates. We then rank them according to the Claim Detection model described in \sectionRef{sec:ranking}. Overall, we obtain $4427$ (ostensible) claims from Wikipedia for the $192$ 
train topics. We fine-tuned GPT-2 on them, and evaluated the results as done for the other datasets (\sectionRef{sec:initial_gen}). Since these data are not manually curated, some of the texts used for fine-tuning are not actual claims.
Nonetheless, human annotators deemed $124/175$ GTs as plausible; average perplexity is $264$, mean prefix ranking accuracy is $0.61$, and average argument quality is $0.75$. These results are comparable to those obtained over the much larger \sbcds{} dataset, suggesting that a good solution to the Claim Retrieval task embodies a good solution to the Claim Generation task.

\section{Further observations}\label{sec:further_obs}

\textbf{What characterizes implausible GTs?}
We considered the $51$ \lxnxgpt{}-FWS test-set GTs which were deemed implausible. More than half seem to contradict common sense, often by connecting pairs of unrelated terms as in the titular \textit{the workweek is the best time to start a family}, for the topic \textit{We should increase the workweek}; or via connecting related terms in an odd manner as in \textit{LGBT adoption is a critical component of a child's life} for the topic \textit{We should legalize LGBT adoption}. Other reasons for implausibility include weird phrasings (e.g., \textit{the housing in public housing is disastrously unaffordable}) and bad grammar (e.g., \textit{that the benefits of the MRT network outweigh its costs}).

\noindent
\textbf{COVID-19 debates.} 
Our pipeline relies heavily on the massive pre-training of GPT-2, that naturally included sentences pertaining -- at least to some extent -- to topics in our dev and test sets. It is therefore interesting to examine the GTs obtained for topics which 
were presumably less abundant 
in the pre-training data.
Hence, while sheltering at home, we have generated $20$ GTs for each of the following two topics: \textit{We should subsidize the COVID-19 drug development} and 
\textit{Coronavirus face masks should be mandatory} 
using the \lxnxgpt{}-FWS model. For the first topic, only $4$ of the $20$ GTs were coherent and relevant, while many of the others talked about HIV, alluded to the opioid crisis, or were outright absurd -- \textit{the use of artificial sweeteners in food should be a crime}. The four ``good" ones were 
of generic 
form, yet some showed an ability to extrapolate to relevant terms, without them being mentioned explicitly in the prefix. For example, in the GT \textit{the COVID-19 vaccine will be a very effective vaccine as compared to other vaccines}, while ``COVID-19" and ``vaccine" are mentioned separately in the prefix (i.e., in the first sentence of the Wikipedia page \textit{COVID-19 drug development}), the term ``COVID-19 vaccine" is not.
For the second topic, $12$ of the GTs are coherent and relevant, presumably because the use of face masks to prevent disease is more general, and may have have been discussed in the pre-training data. It has probably been true of previous airborne viruses that, for example, \textit{the use of face masks is the best way to keep people safe}. Among the irrelevant GTs there is mention of other medical conditions, such as Ebola, diarrhoea and mosquito bites. The full list of GTs for these two topics, as well as $3$ additional ones, are made available as part of this work.

\section{Conclusions}\label{sec:conclusions}

We suggest a claim-generation pipeline, based on a fine-tuned GPT-2 model augmented by framing the topic, and filtered using Claim Detection tools. Results on a diverse set of $96$ new topics demonstrate the merit of our approach. 
As expected, fine tuning on a larger dataset of claims leads to more accurate generation. Yet, the coherency of the dataset also matters; simple merging of datasets of different flavors does not improve generation, and may even hamper it.

 To evaluate the generation models we examined several measures, which roughly estimate how ``good" the generated text is. But since they do so from different perspectives, they are often not consistent with one another \cite{wachsmuth-etal-2017-argumentation}. Here  they were combined heuristically, but future work should explore this more rigorously. 
 

Our work highlights some of the relations between Claim Generation, Claim Retrieval, and Claim Detection. In our pipeline, Claim Detection is used to weed out poorly-generated claims. Further, we show that Claim Retrieval is a sufficient basis -- alongside a powerful language model -- for building a claim generation pipeline; and that Claim Generation may augment Claim Retrieval with additional novel claims.

Here, GPT-2 was used with a ``default" setting. However, there is clearly an interesting trade-off between creativity and coherence, and balancing the two to fit an intended use case -- perhaps even interactively -- which we intend to explore in future research. 

Finally, the claims generated by our pipeline display both subjective opinions and factual assertions. In the latter case, our initial analysis indicates that the generated claims of a factual nature are often, but certainly not always, factually true. Thus, our work highlights a new emerging front in the rapidly expanding area of fact verification -- that of distinguishing valid factual statements from non--valid ones, on top of automatically generated texts. 

\section{Ethical note}
Argument generation has the potential of being misused \cite{solaiman2019release}, as it can potentially allow to automatically generate a variety of false assertions regarding a topic of interest. In addition, GPT-2 text generations have been shown to exhibit different levels of bias towards different demographics \cite{DBLP:conf/emnlp/ShengCNP19}. Nonetheless, the way to address these dangers is for the community to recognize and better understand the properties of such generated texts, and we hope this work provides a step forward in this direction. As, to the best of our knowledge, this is the first work leveraging GPT-2 in the context of argumentation, such work can be used to advance research in the argument generation community, by surfacing issues of such systems. Furthermore, in our setting we allow for arguments to be generated on both sides of the topic, thus if one side is misrepresented, it would be easily uncovered.

\bibliographystyle{acl_natbib}
\bibliography{anthology,main}

\begin{thebibliography}{35}
\expandafter\ifx\csname natexlab\endcsname\relax\def\natexlab#1{#1}\fi

\bibitem[{Ajjour et~al.(2019)Ajjour, Alshomary, Wachsmuth, and
  Stein}]{ajjour-etal-2019-modeling}
Yamen Ajjour, Milad Alshomary, Henning Wachsmuth, and Benno Stein. 2019.
\newblock \href {https://doi.org/10.18653/v1/D19-1290} {Modeling frames in
  argumentation}.
\newblock In \emph{Proceedings of the 2019 Conference on Empirical Methods in
  Natural Language Processing and the 9th International Joint Conference on
  Natural Language Processing (EMNLP-IJCNLP)}, pages 2922--2932, Hong Kong,
  China. Association for Computational Linguistics.

\bibitem[{Anaby-Tavor et~al.(2019)Anaby-Tavor, Carmeli, Goldbraich, Kantor,
  Kour, Shlomov, Tepper, and Zwerdling}]{anabytavor2019data}
Ateret Anaby-Tavor, Boaz Carmeli, Esther Goldbraich, Amir Kantor, George Kour,
  Segev Shlomov, Naama Tepper, and Naama Zwerdling. 2019.
\newblock \href {http://arxiv.org/abs/1911.03118} {Not enough data? deep
  learning to the rescue!}

\bibitem[{Bar-Haim et~al.(2017)Bar-Haim, Bhattacharya, Dinuzzo, Saha, and
  Slonim}]{bar-haim-etal-2017-stance}
Roy Bar-Haim, Indrajit Bhattacharya, Francesco Dinuzzo, Amrita Saha, and Noam
  Slonim. 2017.
\newblock \href {https://www.aclweb.org/anthology/E17-1024} {Stance
  classification of context-dependent claims}.
\newblock In \emph{Proceedings of the 15th Conference of the {E}uropean Chapter
  of the Association for Computational Linguistics: Volume 1, Long Papers},
  pages 251--261, Valencia, Spain. Association for Computational Linguistics.

\bibitem[{Beers and Nagy(2009)}]{beers2009syntactic}
Scott~F Beers and William~E Nagy. 2009.
\newblock Syntactic complexity as a predictor of adolescent writing quality:
  Which measures? which genre?
\newblock \emph{Reading and Writing}, 22(2):185--200.

\bibitem[{Bilu et~al.(2019)Bilu, Gera, Hershcovich, Sznajder, Lahav,
  Moshkowich, Malet, Gavron, and Slonim}]{bilu2019argument}
Yonatan Bilu, Ariel Gera, Daniel Hershcovich, Benjamin Sznajder, Dan Lahav, Guy
  Moshkowich, Anael Malet, Assaf Gavron, and Noam Slonim. 2019.
\newblock Argument invention from first principles.
\newblock In \emph{Proceedings of the 57th Annual Meeting of the Association
  for Computational Linguistics}, pages 1013--1026. Association for
  Computational Linguistics.

\bibitem[{Bilu and Slonim(2016)}]{bilu-slonim-2016-claim}
Yonatan Bilu and Noam Slonim. 2016.
\newblock \href {https://doi.org/10.18653/v1/P16-2085} {Claim synthesis via
  predicate recycling}.
\newblock In \emph{Proceedings of the 54th Annual Meeting of the Association
  for Computational Linguistics (Volume 2: Short Papers)}, pages 525--530,
  Berlin, Germany. Association for Computational Linguistics.

\bibitem[{Boltuzic and Snajder(2014)}]{Boltuzic2014BackUY}
Filip Boltuzic and J.~Snajder. 2014.
\newblock Back up your stance: Recognizing arguments in online discussions.
\newblock In \emph{ArgMining@ACL}.

\bibitem[{Carenini and Moore(2006)}]{carenini2006generating}
Giuseppe Carenini and Johanna~D Moore. 2006.
\newblock Generating and evaluating evaluative arguments.
\newblock \emph{Artificial Intelligence}, 170(11):925--952.

\bibitem[{Cer et~al.(2017)Cer, Diab, Agirre, Lopez-Gazpio, and
  Specia}]{cer2017semeval}
Daniel Cer, Mona Diab, Eneko Agirre, Inigo Lopez-Gazpio, and Lucia Specia.
  2017.
\newblock Semeval-2017 task 1: Semantic textual similarity-multilingual and
  cross-lingual focused evaluation.
\newblock \emph{Eleventh International Workshop on Semantic Evaluations}.

\bibitem[{Devlin et~al.(2018)Devlin, Chang, Lee, and
  Toutanova}]{devlin2018bert}
Jacob Devlin, Ming-Wei Chang, Kenton Lee, and Kristina Toutanova. 2018.
\newblock Bert: Pre-training of deep bidirectional transformers for language
  understanding.
\newblock \emph{arXiv preprint arXiv:1810.04805}.

\bibitem[{Durmus et~al.(2019)Durmus, Ladhak, and
  Cardie}]{durmus-etal-2019-role}
Esin Durmus, Faisal Ladhak, and Claire Cardie. 2019.
\newblock \href {https://doi.org/10.18653/v1/D19-1568} {The role of pragmatic
  and discourse context in determining argument impact}.
\newblock In \emph{Proceedings of the 2019 Conference on Empirical Methods in
  Natural Language Processing and the 9th International Joint Conference on
  Natural Language Processing (EMNLP-IJCNLP)}, pages 5668--5678, Hong Kong,
  China. Association for Computational Linguistics.

\bibitem[{Ein-Dor et~al.(2020)Ein-Dor, Shnarch, Dankin, Halfon, Sznajder, Gera,
  Alzate, Gleize, Choshen, Hou et~al.}]{EinDor2019CorpusWA}
Liat Ein-Dor, Eyal Shnarch, Lena Dankin, Alon Halfon, Benjamin Sznajder, Ariel
  Gera, Carlos Alzate, Martin Gleize, Leshem Choshen, Yufang Hou, et~al. 2020.
\newblock Corpus wide argument mining--a working solution.
\newblock In \emph{Proceedings of the Thirty-Fourth AAAI Conference on
  Artificial Intelligence}.

\bibitem[{El~Baff et~al.(2019)El~Baff, Wachsmuth, Al~Khatib, Stede, and
  Stein}]{el-baff-etal-2019-computational}
Roxanne El~Baff, Henning Wachsmuth, Khalid Al~Khatib, Manfred Stede, and Benno
  Stein. 2019.
\newblock \href {https://doi.org/10.18653/v1/W19-8607} {Computational
  argumentation synthesis as a language modeling task}.
\newblock In \emph{Proceedings of the 12th International Conference on Natural
  Language Generation}, pages 54--64, Tokyo, Japan. Association for
  Computational Linguistics.

\bibitem[{Fan et~al.(2018)Fan, Lewis, and Dauphin}]{fan-etal-2018-hierarchical}
Angela Fan, Mike Lewis, and Yann Dauphin. 2018.
\newblock \href {https://doi.org/10.18653/v1/P18-1082} {Hierarchical neural
  story generation}.
\newblock In \emph{Proceedings of the 56th Annual Meeting of the Association
  for Computational Linguistics (Volume 1: Long Papers)}, pages 889--898,
  Melbourne, Australia. Association for Computational Linguistics.

\bibitem[{Gretz et~al.(2020)Gretz, Friedman, Cohen-Karlik, Toledo, Lahav,
  Aharonov, and Slonim}]{gretz2019largescale}
Shai Gretz, Roni Friedman, Edo Cohen-Karlik, Assaf Toledo, Dan Lahav, Ranit
  Aharonov, and Noam Slonim. 2020.
\newblock A large-scale dataset for argument quality ranking: Construction and
  analysis.
\newblock In \emph{Proceedings of the Thirty-Fourth AAAI Conference on
  Artificial Intelligence}.

\bibitem[{Hidey and McKeown(2019)}]{hidey-mckeown-2019-fixed}
Christopher Hidey and Kathy McKeown. 2019.
\newblock \href {https://doi.org/10.18653/v1/N19-1174} {Fixed that for you:
  Generating contrastive claims with semantic edits}.
\newblock In \emph{Proceedings of the 2019 Conference of the North {A}merican
  Chapter of the Association for Computational Linguistics: Human Language
  Technologies, Volume 1 (Long and Short Papers)}, pages 1756--1767,
  Minneapolis, Minnesota. Association for Computational Linguistics.

\bibitem[{Holtzman et~al.(2019)Holtzman, Buys, Du, Forbes, and
  Choi}]{holtzman2019curious}
Ari Holtzman, Jan Buys, Li~Du, Maxwell Forbes, and Yejin Choi. 2019.
\newblock \href {http://arxiv.org/abs/1904.09751} {The curious case of neural
  text degeneration}.

\bibitem[{Hua et~al.(2019)Hua, Hu, and
  Wang}]{hua-etal-2019-argument-generation}
Xinyu Hua, Zhe Hu, and Lu~Wang. 2019.
\newblock \href {https://doi.org/10.18653/v1/P19-1255} {Argument generation
  with retrieval, planning, and realization}.
\newblock In \emph{Proceedings of the 57th Annual Meeting of the Association
  for Computational Linguistics}, pages 2661--2672, Florence, Italy.
  Association for Computational Linguistics.

\bibitem[{Hua and Wang(2018)}]{hua-wang-2018-neural}
Xinyu Hua and Lu~Wang. 2018.
\newblock \href {https://doi.org/10.18653/v1/P18-1021} {Neural argument
  generation augmented with externally retrieved evidence}.
\newblock In \emph{Proceedings of the 56th Annual Meeting of the Association
  for Computational Linguistics (Volume 1: Long Papers)}, pages 219--230,
  Melbourne, Australia. Association for Computational Linguistics.

\bibitem[{Keskar et~al.(2019)Keskar, McCann, Varshney, Xiong, and
  Socher}]{keskar2019ctrl}
Nitish~Shirish Keskar, Bryan McCann, Lav~R Varshney, Caiming Xiong, and Richard
  Socher. 2019.
\newblock Ctrl: A conditional transformer language model for controllable
  generation.
\newblock \emph{arXiv preprint arXiv:1909.05858}.

\bibitem[{Levy et~al.(2014)Levy, Bilu, Hershcovich, Aharoni, and
  Slonim}]{levy-etal-2014-context}
Ran Levy, Yonatan Bilu, Daniel Hershcovich, Ehud Aharoni, and Noam Slonim.
  2014.
\newblock \href {https://www.aclweb.org/anthology/C14-1141} {Context dependent
  claim detection}.
\newblock In \emph{Proceedings of {COLING} 2014, the 25th International
  Conference on Computational Linguistics: Technical Papers}, pages 1489--1500,
  Dublin, Ireland. Dublin City University and Association for Computational
  Linguistics.

\bibitem[{Orbach et~al.(2019)Orbach, Bilu, Gera, Kantor, Dankin, Lavee,
  Kotlerman, Mirkin, Jacovi, Aharonov, and Slonim}]{orbach-etal-2019-dataset}
Matan Orbach, Yonatan Bilu, Ariel Gera, Yoav Kantor, Lena Dankin, Tamar Lavee,
  Lili Kotlerman, Shachar Mirkin, Michal Jacovi, Ranit Aharonov, and Noam
  Slonim. 2019.
\newblock \href {https://doi.org/10.18653/v1/D19-1561} {A dataset of
  general-purpose rebuttal}.
\newblock In \emph{Proceedings of the 2019 Conference on Empirical Methods in
  Natural Language Processing and the 9th International Joint Conference on
  Natural Language Processing (EMNLP-IJCNLP)}, pages 5591--5601, Hong Kong,
  China. Association for Computational Linguistics.

\bibitem[{Radford et~al.(2019)Radford, Wu, Child, Luan, Amodei, and
  Sutskever}]{radford2019language}
Alec Radford, Jeffrey Wu, Rewon Child, David Luan, Dario Amodei, and Ilya
  Sutskever. 2019.
\newblock Language models are unsupervised multitask learners.
\newblock \emph{OpenAI Blog}, 1(8):9.

\bibitem[{Reisert et~al.(2015)Reisert, Inoue, Okazaki, and
  Inui}]{reisert2015computational}
Paul Reisert, Naoya Inoue, Naoaki Okazaki, and Kentaro Inui. 2015.
\newblock A computational approach for generating toulmin model argumentation.
\newblock In \emph{Proceedings of the 2nd Workshop on Argumentation Mining},
  pages 45--55.

\bibitem[{Rinott et~al.(2015)Rinott, Dankin, Alzate~Perez, Khapra, Aharoni, and
  Slonim}]{rinott-etal-2015-show}
Ruty Rinott, Lena Dankin, Carlos Alzate~Perez, Mitesh~M. Khapra, Ehud Aharoni,
  and Noam Slonim. 2015.
\newblock \href {https://doi.org/10.18653/v1/D15-1050} {Show me your evidence -
  an automatic method for context dependent evidence detection}.
\newblock In \emph{Proceedings of the 2015 Conference on Empirical Methods in
  Natural Language Processing}, pages 440--450, Lisbon, Portugal. Association
  for Computational Linguistics.

\bibitem[{Schiller et~al.(2020)Schiller, Daxenberger, and
  Gurevych}]{schiller2020aspect}
Benjamin Schiller, Johannes Daxenberger, and Iryna Gurevych. 2020.
\newblock Aspect-controlled neural argument generation.
\newblock \emph{arXiv preprint arXiv:2005.00084}.

\bibitem[{Sheng et~al.(2019)Sheng, Chang, Natarajan, and
  Peng}]{DBLP:conf/emnlp/ShengCNP19}
Emily Sheng, Kai{-}Wei Chang, Premkumar Natarajan, and Nanyun Peng. 2019.
\newblock \href {https://doi.org/10.18653/v1/D19-1339} {The woman worked as a
  babysitter: On biases in language generation}.
\newblock In \emph{Proceedings of the 2019 Conference on Empirical Methods in
  Natural Language Processing and the 9th International Joint Conference on
  Natural Language Processing, {EMNLP-IJCNLP} 2019, Hong Kong, China, November
  3-7, 2019}, pages 3405--3410. Association for Computational Linguistics.

\bibitem[{Shnayderman et~al.(2019)Shnayderman, Ein-Dor, Mass, Halfon, Sznajder,
  Spector, Katz, Sheinwald, Aharonov, and Slonim}]{shnayderman2019fast}
Ilya Shnayderman, Liat Ein-Dor, Yosi Mass, Alon Halfon, Benjamin Sznajder,
  Artem Spector, Yoav Katz, Dafna Sheinwald, Ranit Aharonov, and Noam Slonim.
  2019.
\newblock \href {http://arxiv.org/abs/1908.06785} {Fast end-to-end
  wikification}.

\bibitem[{Solaiman et~al.(2019)Solaiman, Brundage, Clark, Askell, Herbert-Voss,
  Wu, Radford, Krueger, Kim, Kreps, McCain, Newhouse, Blazakis, McGuffie, and
  Wang}]{solaiman2019release}
Irene Solaiman, Miles Brundage, Jack Clark, Amanda Askell, Ariel Herbert-Voss,
  Jeff Wu, Alec Radford, Gretchen Krueger, Jong~Wook Kim, Sarah Kreps, Miles
  McCain, Alex Newhouse, Jason Blazakis, Kris McGuffie, and Jasmine Wang. 2019.
\newblock \href {http://arxiv.org/abs/1908.09203} {Release strategies and the
  social impacts of language models}.

\bibitem[{Stab et~al.(2018)Stab, Miller, Schiller, Rai, and
  Gurevych}]{stab-etal-2018-cross}
Christian Stab, Tristan Miller, Benjamin Schiller, Pranav Rai, and Iryna
  Gurevych. 2018.
\newblock \href {https://doi.org/10.18653/v1/D18-1402} {Cross-topic argument
  mining from heterogeneous sources}.
\newblock In \emph{Proceedings of the 2018 Conference on Empirical Methods in
  Natural Language Processing}, pages 3664--3674, Brussels, Belgium.
  Association for Computational Linguistics.

\bibitem[{Wachsmuth et~al.(2017)Wachsmuth, Naderi, Habernal, Hou, Hirst,
  Gurevych, and Stein}]{wachsmuth-etal-2017-argumentation}
Henning Wachsmuth, Nona Naderi, Ivan Habernal, Yufang Hou, Graeme Hirst, Iryna
  Gurevych, and Benno Stein. 2017.
\newblock \href {https://doi.org/10.18653/v1/P17-2039} {Argumentation quality
  assessment: Theory vs. practice}.
\newblock In \emph{Proceedings of the 55th Annual Meeting of the Association
  for Computational Linguistics (Volume 2: Short Papers)}, pages 250--255,
  Vancouver, Canada. Association for Computational Linguistics.

\bibitem[{Wachsmuth et~al.(2018)Wachsmuth, Stede, El~Baff, Al-Khatib,
  Skeppstedt, and Stein}]{wachsmuth-etal-2018-argumentation}
Henning Wachsmuth, Manfred Stede, Roxanne El~Baff, Khalid Al-Khatib, Maria
  Skeppstedt, and Benno Stein. 2018.
\newblock \href {https://www.aclweb.org/anthology/C18-1318} {Argumentation
  synthesis following rhetorical strategies}.
\newblock In \emph{Proceedings of the 27th International Conference on
  Computational Linguistics}, pages 3753--3765, Santa Fe, New Mexico, USA.
  Association for Computational Linguistics.

\bibitem[{Walton and Gordon(2012)}]{walton2012carneades}
Douglas Walton and Thomas~F Gordon. 2012.
\newblock The carneades model of argument invention.
\newblock \emph{Pragmatics \& Cognition}, 20(1):1--31.

\bibitem[{Zellers et~al.(2019)Zellers, Holtzman, Rashkin, Bisk, Farhadi,
  Roesner, and Choi}]{NIPS2019_9106}
Rowan Zellers, Ari Holtzman, Hannah Rashkin, Yonatan Bisk, Ali Farhadi,
  Franziska Roesner, and Yejin Choi. 2019.
\newblock \href
  {http://papers.nips.cc/paper/9106-defending-against-neural-fake-news.pdf}
  {Defending against neural fake news}.
\newblock In H.~Wallach, H.~Larochelle, A.~Beygelzimer, F.~d'Alch\'{e} Buc,
  E.~Fox, and R.~Garnett, editors, \emph{Advances in Neural Information
  Processing Systems 32}, pages 9054--9065. Curran Associates, Inc.

\bibitem[{Zukerman et~al.(1998)Zukerman, McConachy, and
  Korb}]{zukerman1998bayesian}
Ingrid Zukerman, Richard McConachy, and Kevin~B Korb. 1998.
\newblock Bayesian reasoning in an abductive mechanism for argument generation
  and analysis.
\newblock In \emph{AAAI/IAAI}, pages 833--838.

\end{thebibliography}

\appendix

\section{Examples of Labeled GTs}

Tables \ref{table:examplesLxnxTest} and \ref{table:examplesSbcTest} detail GTs generated over the test set by the two models. Each table details two GTs which were labeled implausible, two that were labeled as plausible but lacking stance toward the topic prompt, and two that are coherent GTs for the topic.

\begin{table*}[th]
\scriptsize
\begin{center}
\begin{tabular}{ |p{2.5cm}|p{4cm}|p{1.3cm}|p{5.5cm}|}
\hline
\textbf{Topic} & \textbf{GT} & \textbf{Label} & \textbf{Comment} \\
\hline
We should close landfills & landfills are not a sustainable way to fund the nation's health& implausible& The text is coherent. However, landfills were never intended to provide revenue that will somehow contribute to health funds, which makes this claim implausible.\\ \hline
Criticism of the Iraq War is justified& the Iraq War is a costly and costly mistake & implausible& The odd repetition of the term ``costly" is why this claim is considered implausible.\\ \hline
We should nationalize the health care industry& The health care sector is one of the fastest growth sectors of the economy.& no stance& The claim is praising the effects of the health care industry on the economy, thus suggesting to keep things as they are. However, it could be also interpreted as something that governments could profit from, thus suggesting to support the topic. \\ \hline
We should abandon Valentine's day & The Valentine's Day holiday is one of the most misunderstood and misused holidays. & no stance& The attitude towards Valientine's Day can be understood as negative, as it is misunderstood, or as positive, as something that should be actually getting better respect from the public. In addition, the use of the verb ``misused" is rather \textit{misused} in this context. Either way, there is no clear stance towards abandoning it.  \\ \hline
We should disband the United Nations Security Council & the United Nations Security Council is an essential forum for maintaining the international community's collective efforts to fight terrorism& plausible and has stance &\\ \hline
We should subsidize transportation network companies& the introduction of regional mass transit networks in the country will help boost economic growth, provide enhanced fuel efficiency and reduce the use of private vehicles & plausible and has stance &\\ \hline
 \end{tabular}
 \end{center}
 \caption{Examples of GTs generated by \lxnxgpt{}-FWS on the test set.}
\label{table:examplesLxnxTest}
\end{table*}
\begin{table*}[t]
\scriptsize
\begin{center}
\begin{tabular}{ |p{2.5cm}|p{4cm}|p{1.3cm}|p{5.5cm}|}
\hline
\textbf{Topic} & \textbf{GT} & \textbf{Label} & \textbf{Comment} \\

\hline
We should ban alcohol advertising & alcohol advertising is a form of harassment towards those who may be a heavy drinker/drinker. it goes against the most basic human rights. & implausible& The text is coherent, and for many topics saying that something is a harassment to the point of violating basic human rights is often a plausible argument. But saying that alcohol advertising is a harassment of drinkers doesn't make enough sense to be a plausible argument. \\ \hline
We should abolish competition laws& competition laws are unfair to those who don't have the money to hire a lawyer to get away with murder.& implausible& Here, too, the text is coherent. However, while it's not clear why competition laws make lawyers more expensive (usually competition drives prices down), the argument in implausible because getting away with muder is not something people are likely to argue for.\\ \hline
We should disband the United Nations Security Council & The United Nations system should be reformed to ensure that the council serving as the ultimate check on the world is a more efficient mechanism to deal with global issues. & no stance& Calling for a reform can be taken as an alterntive to disbanding, interpreting the claim as contesting the topic. Alternatively, it can be seen as critical of UNSC, and as milder policy in the same venue as disbanding.\\ \hline
We should increase the use of telemedicine& telemarketing can be a effective means of marketing products and conducting research & no stance& The claim is plausible, but it's not relevant to the topic - it talks about telemarketing instead of telemedicine. This is probably due to telemarketing appearing in the training set.\\ \hline
We should disband the United Nations Security Council & we should not dissolve the united nations security council because it is an important forum to address global issues such as how to deal with unique situations involving nuclear war, natural disasters, and the like.& plausible and has stance & \\ \hline
Flu vaccination should be mandatory & mandatory vaccination is a good thing. it keeps kids safe and ensures that those in the most need of protection are protected. it can help provide a stronger immune system to fight disease and protect them from harmful situations. & plausible and has stance & \\ \hline
 \end{tabular}
 \end{center}
 \caption{Examples of GTs generated by \sbcgpt{} on the test set.}
\label{table:examplesSbcTest}
\end{table*}
\section{Framing Claims}
\label{sec_framing}
In an attempt to frame the GTs,
we append to the topic a short sentence describing an aspect related to the claim, hypothesizing that adding a concrete aspect will guide the generation process in that direction.
We consider the aspects (or frames) appearing $\geq 100$ times in the dataset of \citet{ajjour-etal-2019-modeling}, and manually map each aspect to a related list of Wikipedia pages. Using Wikification, we keep in the training set only claims that reference at least one of these Wikipedia pages. Finally, we manually phrase each aspect as a framing sentence, e.g., \textit{Consider how this relates to the economy} for the \textit{Economy} aspect, and append it to the topic separated by a delimiter.

For evaluation, we generated $15$ GTs per aspect per topic. We compared the results to the \lxnxgpt{} and \sbcgpt{} models, using the same measures as described in the main text. Doing an internal manual assessment of a sample of $40$ GTs for each model, we found that adding aspect context did not improve the plausibility and relevance of GTs, not even when introducing heuristics to detect aspects that are more relevant to the topic. A possible explanation for this is that the selection of appropriate aspects should be handled more carefully (e.g., as in \citet{schiller2020aspect}). Such an approach is beyond the scope of this work, and we leave it for future work.
\section{Using Claim Detection to Rank GTs}

When constructing our pipeline, we examined $3$ models for ranking GTs according to their coherence and relevance, concluding that the Claim Detection (CD) model is most successful. This model is obtained by fine-tuning BERT on a similar dataset to what was used to fine-tune \lxnxgpt{} (the main difference is that the data used to fine-tune BERT included also negative examples from the same corpus), thus reminiscent of bootstrapping. Indeed, this method of using a classifier fine-tuned on the same data as GPT-2 to filter generated samples has already proven to be effective in the context of augmenting low-resource datasets with generated texts \cite{anabytavor2019data}.
\section{Lists of topics}
\subsection{List of dev set topics}
We should legalize doping in sport                               \\ 
We should protect endangered species                             \\ 
We should legalize insider trading                               \\ 
We should lower the drinking age                                 \\ 
We should abolish temporary employment                           \\ 
We should ban free newspapers                                    \\ 
We should abolish the US Electoral College                       \\ 
We should ban lotteries                                          \\ 
We should legalize ivory trade                                   \\ 
We should further exploit green technology                       \\ 
We should ban abortions                                          \\ 
We should further exploit geothermal energy                      \\ 
We should raise the retirement age                               \\ 
We should ban alternative medicine                               \\ 
We should subsidize public service broadcasters                  \\ 
We should abolish term limits                                    \\ 
We should abandon Gmail                                          \\ 
We should not subsidize single parents                           \\ 
We should introduce school vouchers                              \\ 
Prenatal diagnosis should be mandatory                           \\ 
We should prohibit tower blocks                                  \\ 
We should increase airport racial profiling in the United States \\ 
We should increase international volunteering                    \\ 
We should subsidize the human mission to Mars                    \\ 
The use of AI should be abandoned                                \\ 
We should fight for Palestinian independence                     \\ 
We should further exploit natural gas                            \\ 
We should abandon democracy                                      \\ 
We should ban fishing                                            \\ 
We should ban gratuities                                         \\ 
We should increase government regulation                         \\ 
Community service should be mandatory                            \\ 
We should further exploit solar energy                           \\ 
Tattoos should be banned                                         \\ 
We should support a phase-out of lightweight plastic bags       \\ 

\subsection{List of test set topics}

We should end the use of solitary confinement         \\ 
We should disband the United Nations Security Council \\ 
We should end the use of mass surveillance            \\ 
Child labor should be legalized                       \\ 
We should cancel the pledge of allegiance to the flag \\ 
We should ban multi-level marketing                   \\ 
We should adopt environmental justice                 \\ 
We should ban media conglomerates                     \\ 
We should end the use of traffic enforcement cameras  \\ 
We should introduce a national identity card          \\ 
We should subsidize transportation network companies  \\ 
We should ban burqas                                  \\ 
We should ban conversion therapy                      \\ 
We should introduce the alternative vote              \\ 
Force-feeding should be banned                        \\ 
We should abandon tabloid journalism                  \\ 
We should legalize LGBT adoption                      \\ 
We should abandon Twitter                             \\ 
We should abandon chain stores                        \\ 
We should further exploit mixed-use development       \\ 
We should subsidize open access journals              \\ 
We should end child benefits                          \\ 
We should increase the use of telemedicine            \\ 
We should abandon the sexual revolution               \\ 
We should adopt polyamory                             \\ 
We should end the use of bailouts                     \\ 
Begging should be banned                              \\ 
We should adopt catholicism                           \\ 
We should abolish credit scores                       \\ 
We should fight environmental degradation             \\ 
We should increase environmental protection           \\ 
Flu vaccination should be mandatory                   \\ 
We should close landfills                             \\ 
We should further exploit filibusters                 \\ 
Minority groups should be protected                   \\ 
\subsection{Annotation Task Guidelines}

\begin{figure*}[t]
\caption{Guidelines for the plausibility annotation task.}
\centering
\includegraphics[scale=0.7]{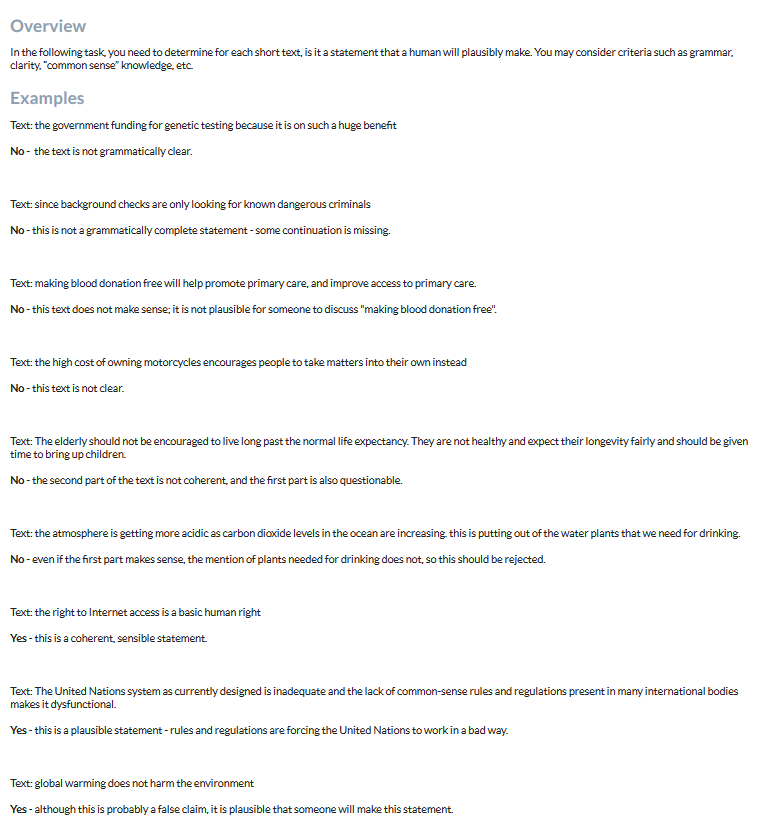}
\label{fig:plausibility}
\end{figure*}

\begin{figure*}[t]
\caption{Example of a factual annotation.}
\centering
\includegraphics[]{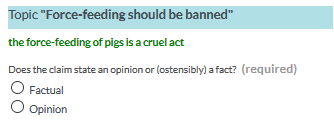}
\label{fig:factualExample}
\end{figure*}

Figures \ref{fig:plausibility}-\ref{fig:factualExample} present the guidelines for the plausibility, stance and factual vs. opinion annotation tasks, as appearing in the Appen crowd-sourcing platform.

\begin{figure*}[t]
\caption{Example of a plausibility annotation.}
\centering
\includegraphics[]{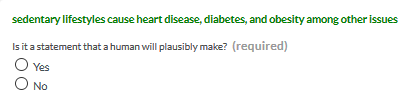}
\end{figure*}

\begin{figure*}[t]
\caption{Guidelines for the stance annotation task.}
\centering
\includegraphics[scale=0.7]{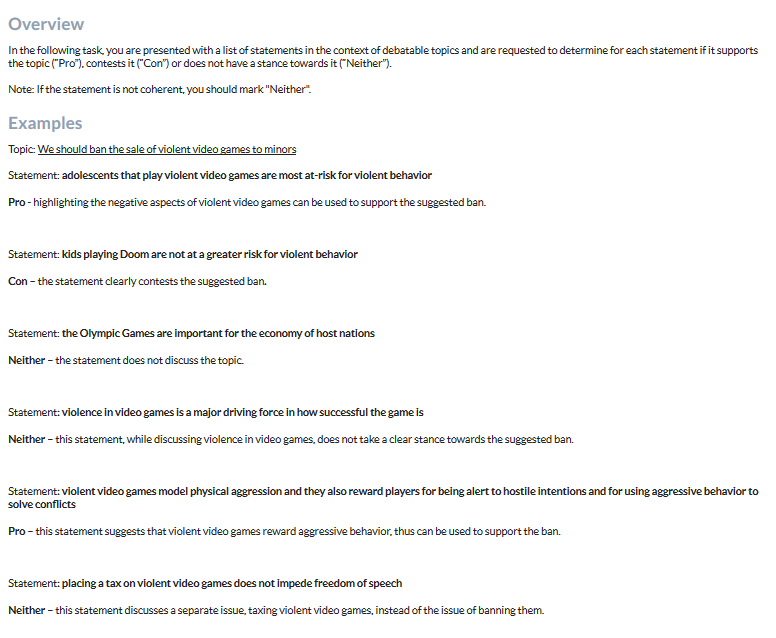}
\end{figure*}

\begin{figure*}[t]
\caption{Example of a stance annotation.}
\centering
\includegraphics[]{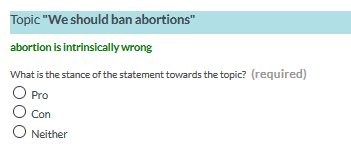}
\end{figure*}

\begin{figure*}[t]
\caption{Guidelines for the factual vs. opinion annotation task.}
\centering
\includegraphics[scale=0.5]{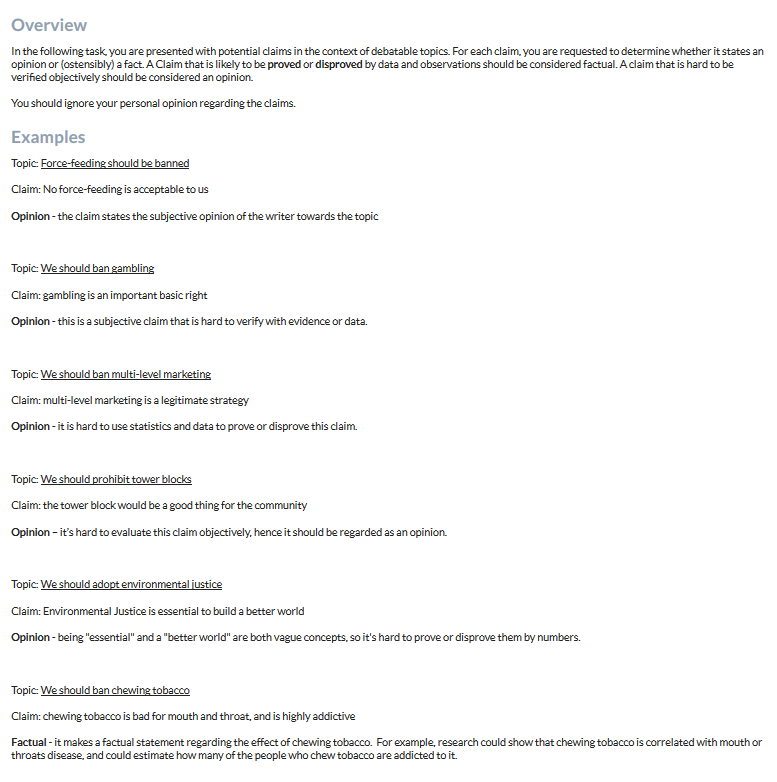}
\includegraphics[scale=0.5]{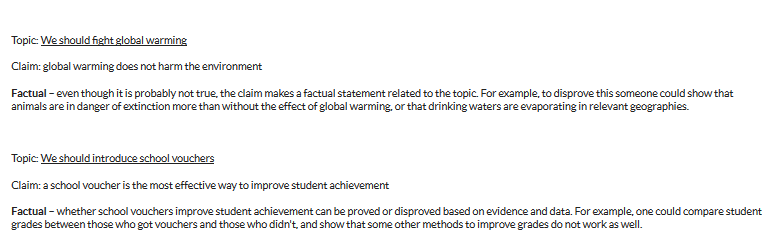}
\end{figure*}

\end{document}